\documentclass[lettersize,journal]{IEEEtran}
\usepackage{amsmath,amsfonts}
\usepackage{hyperref} 
\usepackage{array}
\usepackage[caption=false,font=normalsize,labelfont=sf,textfont=sf]{subfig}
\usepackage{textcomp}
\usepackage{stfloats}
\usepackage{url}
\usepackage{verbatim}
\usepackage{cite}
\usepackage{multirow}
\usepackage[linesnumbered,ruled,vlined,ruled]{algorithm2e}
\usepackage{booktabs}
\usepackage{color}
\usepackage{amsmath}
\usepackage{longtable}
\usepackage{graphicx}
\usepackage{bbding}
\usepackage{caption,subcaption}
\usepackage{threeparttable}
\usepackage{tabularx}
\newcolumntype{R}[1]{>{\raggedleft\arraybackslash}p{#1}}
\newcolumntype{L}[1]{>{\raggedright\arraybackslash}p{#1}}
\newcolumntype{C}[1]{>{\centering\arraybackslash}p{#1}}
\newcolumntype{Y}{>{\raggedright\arraybackslash}X}
\newcolumntype{Z}{>{\raggedleft\arraybackslash}X}
\newcolumntype{A}{>{\centering\arraybackslash}X}
\usepackage{tikz}
\usetikzlibrary{shapes, arrows, positioning}
\tikzstyle{startstop} = [rectangle, rounded corners, minimum width=3cm, minimum height=1cm, text centered, draw=black, fill=red!30]
\tikzstyle{io} = [trapezium, trapezium left angle=70, trapezium right angle=110, minimum width=3cm, minimum height=1cm, text centered, draw=black, fill=blue!30]
\tikzstyle{process} = [rectangle, minimum width=3cm, minimum height=1cm, text centered, draw=black, fill=orange!30]
\tikzstyle{arrow} = [thick,->,>=stealth]
\usepackage[bottom=1.82cm,top=2.01cm,left=1.69cm,right=1.69cm]{geometry}
\begin{document}

\title{SSF-PAN: Semantic Scene Flow-Based Perception for Autonomous Navigation in Traffic Scenarios}

\author{Yinqi Chen, Meiying Zhang, Qi Hao, Guang Zhou % <-this % stops a space
\thanks{Yinqi Chen and Meiying Zhang are co-first authors; Corresponding author: Guang Zhou (email: maxwell@deeproute.ai) and Qi Hao (email: hao.q@sustech.edu.cn)}
\thanks{Yinqi Chen, Meiying Zhang, Qi Hao are with the Research Institute of Trustworthy Autonomous Systems, Southern University of Science and Technology, Shenzhen 518055, China; Guang Zhou is with Operation Department, Shenzhen Deeproute.ai Co.,Ltd, Shenzhen, China.}
\thanks{This work is jointly supported in part by the Hetao Shenzhen-HongKong Science and Technology Innovation Cooeration Zone (HZQB-KCZYZ-2021055), the National Natural Science Foundation of China (62261160654), the Shenzhen Fundamental Research Program (JCYJ20220818103006012, KJZD20231023092600001), and the Shenzhen Key Laboratory of Robotics and Computer Vision (ZDSYS20220330160557001).
}% <-this % stops a space

}



\maketitle
\thispagestyle{empty}
\begin{abstract}
Vehicle detection and localization in complex traffic scenarios pose significant challenges due to the interference of moving objects. 
Traditional methods often rely on outlier exclusions or semantic segmentations, which suffer from low computational efficiency and accuracy. 
The proposed SSF-PAN can achieve the functionalities of LiDAR point cloud based object detection/localization and SLAM (Simultaneous Localization and Mapping) 
with high computational efficiency and accuracy, enabling map-free navigation frameworks.
The novelty of this work is threefold: 
1) developing a neural network which can achieve segmentation among static and dynamic objects within the scene flows
with different motion features, that is, semantic scene flow (SSF); 
2) developing an iterative framework which can further optimize the quality of input scene flows and output segmentation results; 
3) developing a scene flow-based navigation platform which can test the performance of the SSF perception system in the simulation environment. 
The proposed SSF-PAN method is validated using the SUScape-CARLA \footnote{\url{https://suscape.net/datasets/sceneflow}} and the KITTI \cite{geiger2013vision} datasets, as well as on the CARLA simulator.
Experimental results demonstrate that the proposed approach outperforms traditional methods in terms of scene flow computation accuracy, moving object detection accuracy, computational efficiency, and autonomous navigation effectiveness.
\end{abstract}

\begin{IEEEkeywords}
Semantic Scene Flow, SLAM, Moving Object Detection, Autonomous Navigation.
\end{IEEEkeywords}

\section{Introduction}

\begin{figure}
\centering
\includegraphics[width=0.9\linewidth]{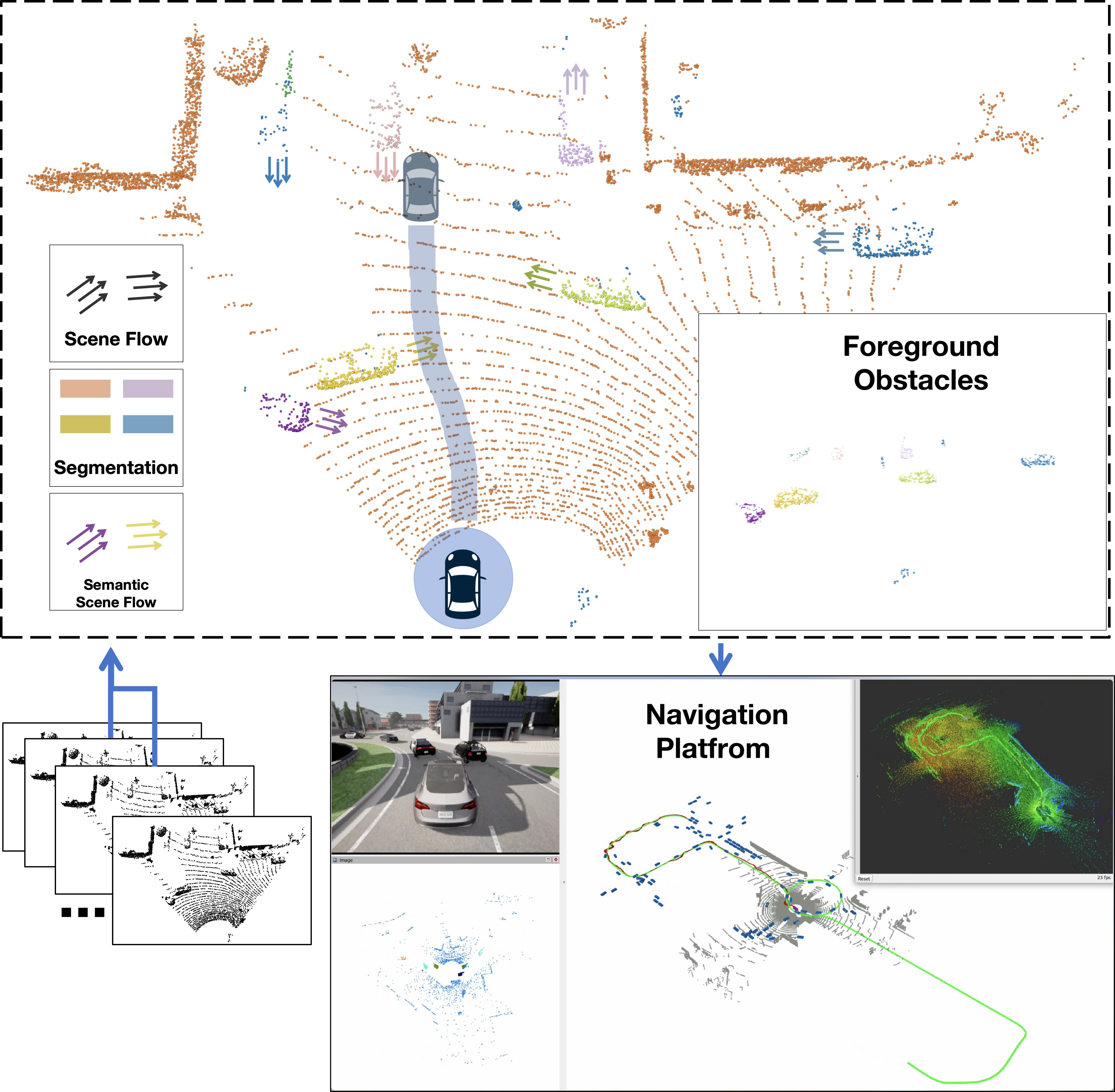}
\caption{An illustration of SSF estimation for autonomous navigation with  dynamic and static object classification as well as moving object instance segmentation.}
\label{fig:introduction}
\vspace{-0.5cm} 
\end{figure}

\IEEEPARstart{A}{s} intelligent transportation systems and autonomous driving technologies advance, 
there is an increasing need for perception systems that can effectively help understand and navigate through complex traffic scenarios. 
Traditional navigation frameworks need Simultaneous Localization and Mapping (SLAM) techniques which often struggle with high computational complexities;
they also need an additional moving object detection module to avoid obstacles.
Recently, scene flow technologies have been developed to achieve map-free perception functionalities,
which estimate the point-wise correlation between two adjacent frames of point clouds and can be used for ego-vehicle localization\cite{alcantarilla2012combining} and instance segmentation\cite{NEURIPS2022_c6e38569}. 
Despite progresses in scene flow based perception research, several challenges remain in handling complex traffic scenarios:
\begin{itemize}
\item \textbf{High-Quality Semantic Segmentation}: 
Many methods exploit semantic information to identify static environments and dynamic objects, but their effectiveness is limited by the quality of object feature extraction and classification\cite{bescos2018dynaslam,yu2018ds}. 
Usually, these methods do not fully utilize motion information, especially for complex dynamic traffic scenes where occlusions frequent happen. 
\item \textbf{Accurate Scene Flow Estimation}: 
Scene flow methodologies generally fall into two categories: 
(1) sparse point cloud samples\cite{8953876,wang2023active,2019HPLFlowNet} and (2) dense point cloud samples\cite{2021Weakly,vedder2023zeroflow,jund2021scalable}. 
The former is advantageous in high computational efficiency and suitable for real-time navigation applications, but its estimation accuracy is challenged by frame-to-frame measurement inconsistency, object occlusions, and low point cloud density.
\item \textbf{Comprehensive Navigation Testing Platform}: 
Existing systems often do not fully address the specific requirements for scene flow and navigation performance in dynamic environments. SSF-PAN provides a scene flow-based navigation platform testable in the CARLA simulator, ensuring the perception system's accuracy and dependability in intricate traffic scenarios. 
Scene flow based navigation systems are advantageous in no need of the high-definition maps for ego-vehicle localization.
A comprehensive navigation testing platform requires the development of a set of scene flow based perception modules for ego-vehicle localization and moving object detection, as well as prediction, decision-making, and motion planning modules\cite{leon2019review}. 
\end{itemize} 

The proposed SSF-PAN system addresses these issues by utilizing the semantic information of scene flows, 
enhancing performance in pose estimation, odometry, and perception robustness in complex traffic scenarios. 
By recursively optimizing the quality of semantic segmentation and the accuracy of scene flow estimation, 
SSF-PAN can achieve higher quality of self-localization and moving object detection, as illustrated in Fig.~\ref{fig:introduction}, leading to more reliable navigation systems.
The main contributions of this work include:
\begin{itemize}
\item \textbf{Advanced Scene Flow-Based Point Cloud Segmentation}: 
We develop a novel neural network which utilizes scene flow information to effectively segment point clouds, distinguishing between static environment and dynamic objects with various motion features. This enhances the understanding and differentiation of object types and their movements in complex traffic scenarios.
\item \textbf{Refined Iterative Optimization Framework}: 
We develop an iterative framework that continuously refines both the input scene flow data and the segmentation outputs. 
This iterative process improves the accuracy and robustness of the system and generates higher quality data throughout the perception pipeline.
\item \textbf{Robust Map-Free Navigation Platform}: 
We develop a comprehensive navigation testing platform based on scene flow data, capable of evaluating the performance of the
SSF perception system within various simulated environments. 
This platform supports robust autonomous navigation without the need of pre-built high definition maps and adapts well to many traffic conditions.
\end{itemize}

\section{RELATED WORK}
% Many autonomous driving and intelligent transportation systems have leveraged scene flow and semantic information to enhance perception performance\cite{baur2021slim,huang2021multibodysync,thomas2021self,yi2018deep,golyanik2017multiframe,vogel2013piecewise}. However, these methods still face challenges in computational complexity and estimation accuracy.

\subsection{Scene Flow based Motion Segmentation}
Estimating 3D motion vectors (scene flow) in point cloud processing effectively aids semantic segmentation.
Recent learning-based approaches use scene flow to perform point-wise segmentation, 
addressing challenges such as data sparsity and the lack of mature representations.
However, current methods\cite{baur2021slim,huang2021multibodysync,yi2018deep} often require ground truth for supervision or are limited to simple foreground background segmentation. 
For instance, Thomas et al. \cite{thomas2021self} trained networks using multiple recordings to classify static, slow-moving, and fast-moving objects but struggled with accurate dynamic object segmentation. 
The OGC\cite{NEURIPS2022_c6e38569} offers self-supervised segmentation but fails to effectively identify background points. 
In contrast, our SSF-PAN framework addresses these limitations by employing self-supervised learning with scene flow data for motion segmentation to accurately segment both background points and coarse-grained dynamic objects,
thereby improving the precision of motion information extraction in traffic scenarios.
\subsection{Scene Flow Optimization} 
To enhance scene flow estimation, numerous methodologies leverage supplementary auxiliary information. 
Some methods \cite{golyanik2017multiframe,vogel2013piecewise} employed rigidity constraints on RGB-D frames and piecewise rigid planar modeling to refine scene flow.
Dewan et al.\cite{dewan2016rigid} applied local geometric consistency on a factor graph using 3D descriptors, while GraphFlow\cite{abu2015graphflow} focused on large-motion scenarios, albeit reliant on sparse keypoints. 
Data-driven methods \cite{ma2019deep}, fused depth and flow estimations but necessitated instance segmentation.  
Concurrently, PointFlowNet\cite{behl2019pointflownet} amalgamated 3D scene flow prediction with PointNet++\cite{qi2017pointnet++} based flow estimators. 
Although certain self-supervised learning approaches\cite{hassani2019unsupervised, thabet2020self, zhang2021self} have mitigated the dependency on annotated datasets, they often still demand pre-training or additional data processing. 
In contrast, our SSF-PAN framework employs scene flow derived from consecutive point clouds, iteratively refining point cloud segmentation information through a self-supervised learning approach. 
This methodology removes the need for supplementary annotated data and extraneous information, resulting in superior accuracy and efficiency within complex traffic scenarios.
\subsection{Map-Free Navigation in Dynamic Environments}
Recent technologies integrate scene flow estimation and semantic segmentation to enhance navigation accuracy and decision-making\cite{vedder2023zeroflow,jund2021scalable}. 
However, these methods often demand substantial computational resources and large datasets. 
In addition, various map-free approaches have been proposed to reduce the dependency on high-definition maps.
Image-based methods \cite{broggi1995vision,aly2008real,seo2014detection}suffer from poor generalization across different lighting conditions, scene dynamics, and infrastructural changes. 
LiDAR-based approaches \cite{ort2018autonomous} offer greater robustness but typically rely on specific local structures. 
End-to-end learning methods \cite{amini2019variational,bojarski2016end,codevilla2018end}do not need high-definition maps but are highly data-intensive and require significant computational resources\cite{kendall2019learning}.
Currently, there are no map-free navigation systems which utilize scene flow data for real-time applications.
Our navigation testing platform enables real-time adaptation to dynamic traffic conditions, enhancing both flexibility and efficiency. 
The closed-loop testing system within SSF-PAN ensures robust and scalable navigation, setting a new benchmark for map-free systems.
\begin{figure*}[!t]
  \centering  
  % \vspace{-1.0cm} 
  \includegraphics[width=0.9\linewidth]{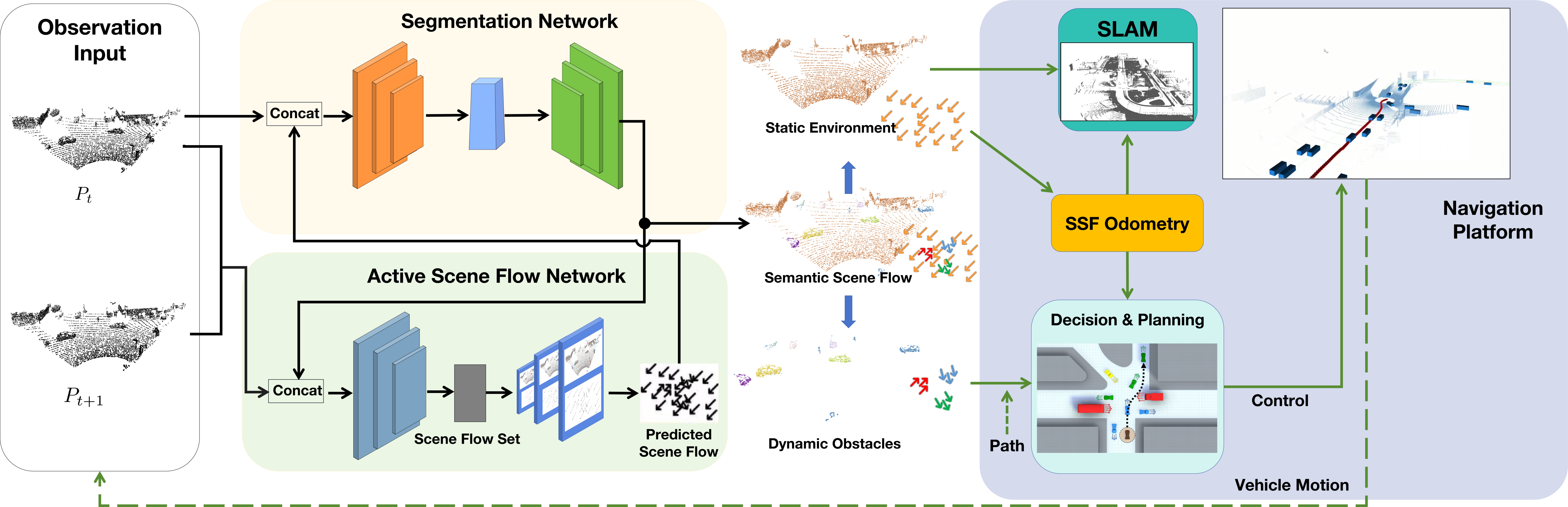}
  \caption{An illustration of the SSF-PAN system diagram.}
  \label{fig:system}
  \vspace{-0.5cm} 
\end{figure*}

\section{System setup and problem statement}
\subsection{System Setup}
This system comprises two main components: a neural network for semantic scene flow estimation and a navigation platform based on this scene flow, as shown in Fig. \ref{fig:system}. 

Two sequential point cloud frames, $P_t$ and $P_{t+1}$, are taken as the input of the system, where $P_t$ is combined with a preliminary semantic mask, to predict the scene flow via \textit{active scene flow} (ASF)\cite{wang2023active} network. Subsequently, the predicted scene flow is concatenated with $P_t$ and processed by the segmentation network to generate the semantic scene flow. This process is iteratively repeated within the mutual promotion network, using the updated mask to refine the input until both the scene flow prediction and segmentation results converge. 
Through the iterative refinement, the networks accurately recognize between static environments and dynamic objects.

After the scene flow estimation and motion segmentation stabilize, the dynamic objects can be regarded as obstacles in navigation planning. By incorporating the navigation path into the comprehensive map, the scene flow data aids in estimating the velocity of each obstacle. The planning algorithm (\textit{e.g.}, RDA planner \cite{10036019}) then uses these information to execute obstacle avoidance planning effectively. The point cloud and scene flow of static environment can be used to compute the transformation matrix for odometry or SLAM.

\subsection{Problem Statement}
Therefore, we focus on solving the following problems:
\begin{enumerate}
\item How to design effective constraints and representations for accurate segmentation of point clouds into static and dynamic objects, leveraging scene flow for enhanced motion segmentation?
\item How to iteratively refine scene flows and segmentation outputs to mutually improve their accuracy, enhancing overall scene flow estimation and motion segmentation?
\item How to integrate SSF into navigation modules for improved pose estimation and obstacle avoidance, enhancing autonomous navigation in traffic scenarios?
\end{enumerate}

\section{Proposed Methods}
\subsection{Scene Flow-based Network for Motion Segmentation}
\begin{figure*}[!h]
  \centering 
  % \vspace{-0.8cm} 
  \includegraphics[width=0.9\linewidth]{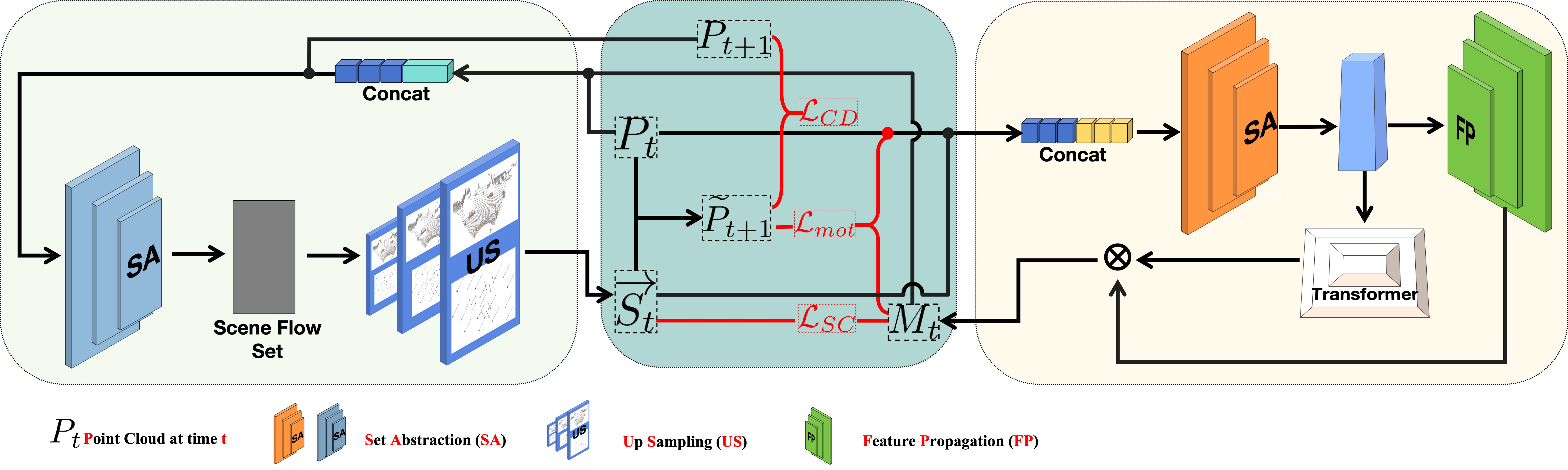}
  \caption{An illustration of the SSF module, which includes two parts: scene flow estimation and motion segmentation.}
  \label{fig:lossfuntion}
  \vspace{-0.5cm} 
\end{figure*}
As shown in Fig. \ref{fig:lossfuntion}, this method begins by acquiring two consecutive frames of point clouds and a initial motion mask. The neural network of semantic scene flow is primarily based on modifications to the ASF and OGC networks. These point clouds are processed using ASF, which leverages spatio-temporal feature consistency to estimate the initial scene flow, $\overrightarrow{S_t}$. Then, it is used as an additional feature input for the segmentation network. Given the point cloud $P_t$ and its corresponding scene flow $\overrightarrow{S_t}$, they are processed to perform motion segmentation, generating a new mask $M_t$ that identifies dynamic objects and the static environment. 
The motion segmentation in our approach is achieved by minimizing the total loss function $\mathcal{L}^{*}$, that is, a combination of motion loss $\mathcal{L}_{mot}$, semantic scene flow consistency loss $\mathcal{L}_{SC}$, and per-cluster rigidity loss using Chamfer distance (CD) $\mathcal{L}_{CD}$. 
\begin{equation}
\begin{aligned}
\mathcal{L}^{*} = \min_{\mathcal{L}} \mathcal{L}_{mot} + \mathcal{L}_{SC} + \mathcal{L}_{CD}
\end{aligned}
\end{equation}
Here, we introduce the following self-supervised components to satisfy the geometry consistency between the $t^{th}$ frame $P_t$ and its predicted next frame $\widetilde{P}_{t+1}$, where $\widetilde{P}_{t+1}=P_t + \overrightarrow{S_t}$.

\textbf{Rigid Motion Consistency Loss $\mathcal{L}_{mot}$:}
    This loss ensures the consistency of rigid body motion between two frames. It is computed by estimating the transformation matrix $T_k \in \mathbb{R}^{4 \times 4}$ ($k=[1,\cdots, K]$) for each object using the weighted-Kabsch algorithm \cite{kabsch1976solution}. The loss minimizes the discrepancy between the transformed points and the estimated scene flow, thereby promoting accurate dynamic motion segmentation.
    \begin{equation}
    \mathcal{L}_{mot} = \frac{1}{K} \sum_{k=1}^{K} \left\| T_k \cdot (P_t \odot M_t^{k}) - (\widetilde{P}_{t+1} \odot M_t^{k} )\right\|_2
    \end{equation}
    where $\odot$ is element-wise product.

\textbf{Semantic Scene Flow Consistency Loss $\mathcal{L}_{SC}$:}
    Enforces consistency of scene flow estimation for all points within the same object,penalizing discrepancies and ensuring uniform motion characteristics:
% \begin{equation}
% N_k = \sum_{i=1}^{N}\delta(M_t,i), \quad \delta(M_t,i)=
% \left\{
% \begin{array}{ll}
% 1, M_{t,i} = k \\
% 0, M_{t,i} \neq k
% \end{array}
% \right.
% \end{equation}
\begin{equation}
\mathcal{L}_{SC} = \frac{1}{K \cdot N_k}\sum_{k=1}^{K} \sum_{i=1}^{N_k} \left( \overrightarrow{S^{k}_{t, i}} - \frac{1}{N_k} \sum_{j=1}^{N_k} \overrightarrow{S^{k}_{t, j}} \right)^2
\end{equation}
where $N_k = \sum_{i=1}^{N}\delta(M_t,i)$, $\delta(M_t,i) = 1$ when $M_{t,i} = k$ and $\delta(M_t,i) = 0$ otherwise, \(\overrightarrow{S^{k}_{t, i}}\) represents the \(i^{\text{th}}\) 
element of $\overrightarrow{S^{k}_{t}}$ and $\overrightarrow{S^{k}_{t}}=\overrightarrow{S_{t}} \odot M_t^k$.

\textbf{Rigidity Chamfer Distance Loss $\mathcal{L}_{CD}$:}
    Maintains spatial consistency within each cluster by minimizing the Chamfer distance between the predicted point and the scanned one at the $(t+1)^{th}$ frame:
    \begin{equation}
    \mathcal{L}_{CD} = \sum_{x \in P_{t+1}} \min_{y \in \widetilde{P}_{t+1}} ||x - y||_2 + \sum_{y \in \widetilde{P}_{t+1}} \min_{x \in P_{t+1}} ||x - y||_2
    \end{equation}

\subsection{Iterative Optimization for Scene Flow and Segmentation}
% \begin{algorithm}[!t]
%   \caption{Iterative Optimization for Scene Flow and Motion Segmentation}
%   \label{algo:SceneFlowOptimization}
%   \SetKwInOut{Input}{Input}
%   \SetKwInOut{Output}{Output}
%   \SetKw{And}{and}
  
%   \Input{
%     $P_t$: Initial point cloud, $M_t^{(0)}$: Initial mask\\
%     $N_{iter}$: Maximum iterations, $\epsilon$: Convergence threshold
%   }
%   \Output{
%     $\overrightarrow{S_t}^{(i)}$: Final scene flow, $M_t^{(i)}$: Final mask
%   }
%   \BlankLine
%   $\Delta_{total} \gets \infty$\\
%   $i \gets 0$\\
  
%   \While{$i < N_{iter}$ \And $\Delta_{total} > \epsilon$}{
%     $\overrightarrow{S_t}^{(i)} \gets$ estimate scene flow based on $(P_t, M_t^{(i)})$\\
%     $M_t^{(i+1)} \gets$ predict motion mask based on $(P_t, \overrightarrow{S_t}^{(i)})$\\
    
%     Check scene flow and mask convergence $\Delta_{total}$ by Eq. (\ref{eq:total})\\
    
%     $i \gets i + 1$\\
%     \textbf{end while}
%   }
%   Identify $M_t^{(s)}$ and $M_t^{(d)}$ from $M_t^{(i)}$ by Eq. (\ref{eq:quatity}) and Eq. (\ref{eq:velocity})\\
%   \KwRet{$\overrightarrow{S_t}^{(i)}$, $M_t^{(i)}$}
% \end{algorithm}
The ASF network \cite{wang2023active} takes the point cloud $P_t$ and an initialized segmentation mask $M_t^{(0)}$ as inputs. This setup allows the network to iteratively refine the scene flow estimation and dynamic \& static segmentation.
During each iteration, the latest segmentation mask $M_t$ is concatenated with the point cloud $P_t$ and passed through the network. The process involves \textit{set abstraction} (SA) feature extraction, \textit{scene flow set}, and \textit{up sampling} (US) modules, which compute a coarse scene flow, $\overrightarrow{S_t}^{(0)}$. This scene flow is utilized for initial motion segmentation, resulting in an updated mask, $M_t$.
The iterative refinement process continues until the change $\Delta_{total}$ in the scene flow estimation and motion segmentation converges to a threshold $\epsilon$, initialized to $10^{-3}$,
\begin{equation}
\label{eq:total}
\Delta_{total} = \alpha \cdot \left\| \overrightarrow{S_t}^{(i)} - \overrightarrow{S_t}^{(i-1)} \right\|_2 + \beta \cdot \left\| M_t^{(i)} - M_t^{(i-1)} \right\|_2
\end{equation}
where $\alpha$ and $\beta$ are weights that balance the contributions of the scene flow and mask convergence.

This ensures that the scene flow and segmentation results stabilize, leading to improved accuracy and robustness in dynamic scene understanding. 
% The iterative optimization is shown in Algorithm \ref{algo:SceneFlowOptimization}.

The system employs two strategies to classify points as static environment with mask $M_t^{(s)}$ and $(k-1)$ dynamic obstacles with mask $M_t^{(d)}$, that is, $M_t = \{M_t^{(s)}, M_t^{(d)}\}$ :

\textbf{Quantity-based Classification}: Based on the assumption that dynamic objects are fewer than static background elements in traffic scenarios, it is simple to determine the static set $M_t^{(s)}$ by statistical quantity for each cluster. The maximum cluster size is considered as the static part:
    \begin{equation}
    \label{eq:quatity}
    M_t^{(s)} = \{M_{t,i} \mid M_{t,i} = \arg\max_{k} N_k \}
    \end{equation}
    
\textbf{Velocity-based Classification}: If the size variance is minimal within a threshold, the system relies on velocity to classify points. It calculates the average velocity $\overline{V}_t^k$ for each cluster:
    \begin{equation}
    \label{eq:avg_V}
    \overline{V}_t^k = \frac{1}{N_k \cdot \Delta t} \sum_{i=1}^{N_k} \left\|\overrightarrow{S^{k}_{t, i}}\right\|
    \end{equation}
     where $\Delta t$ is the time interval between two frames. Then the object whose velocity is close to that of the ego-vehicle, $V_{ego}$, is identified as a static object:
    \begin{equation}
    \label{eq:velocity}
    M_t^{(s)} = \{M_{t,i} \mid \left|\overline{V}_t^k - V_{ego} \right| < \theta\}.
    \end{equation}

\subsection{Scene Flow-Based Navigation Platform}
\begin{figure}
  \centering 
   % \vspace{-0.5cm} 
   \includegraphics[width=0.95\linewidth]{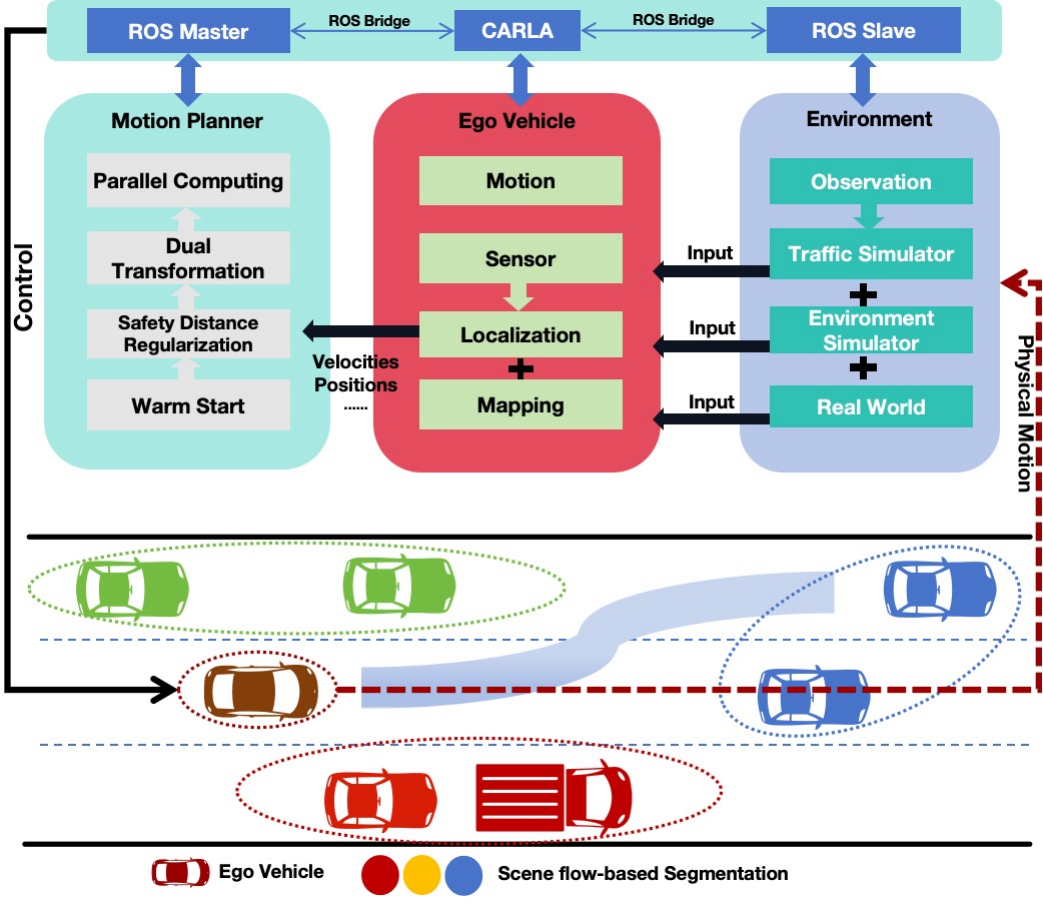}
  \caption{An illustration of the SSF navigation platform in CARLA.}
  \label{fig:testframework}
  \vspace{-0.5cm} 
\end{figure}
\begin{figure}
  \centering 
   % \vspace{-0.5cm} 
   \includegraphics[width=0.8\linewidth]{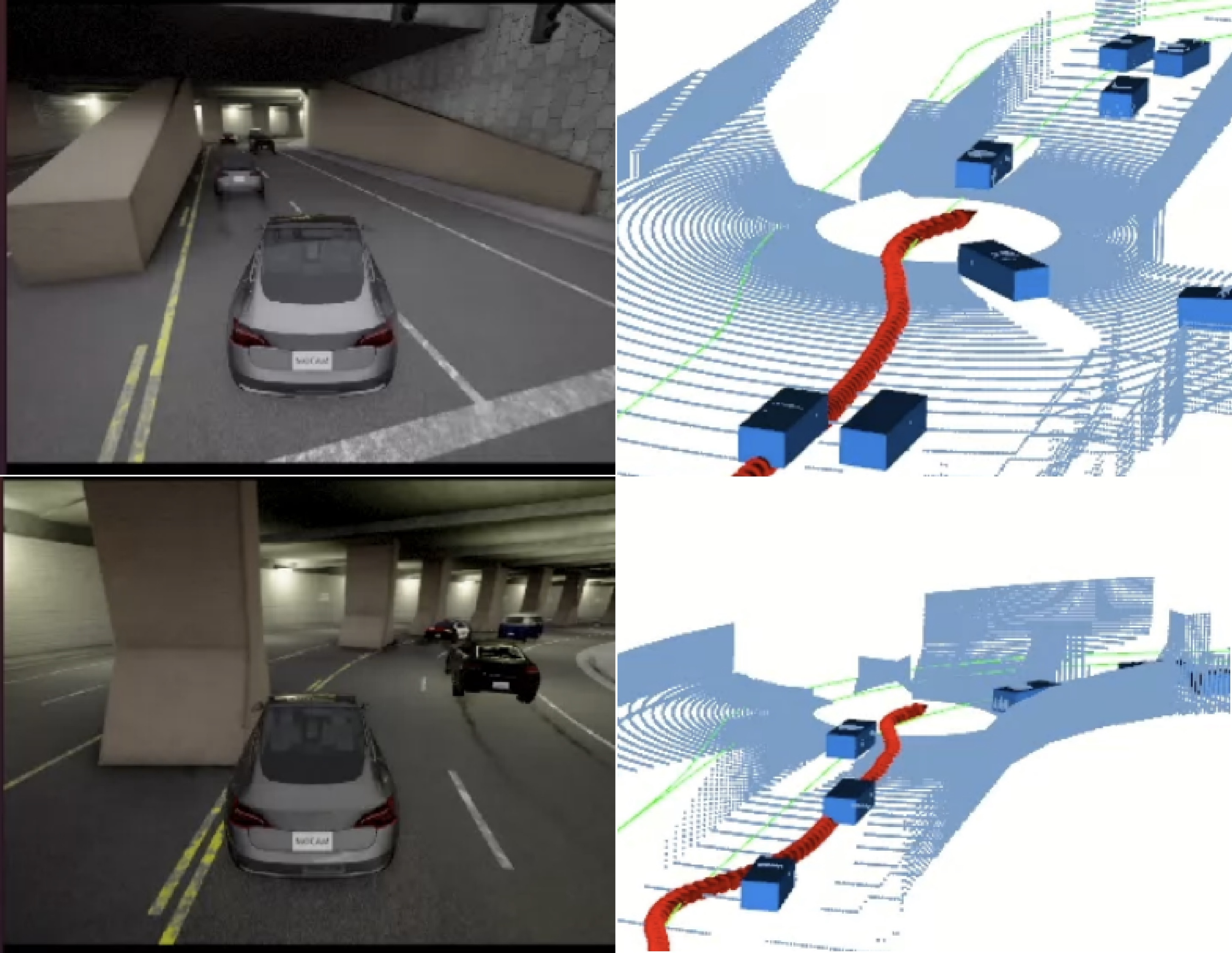}
  \caption{A snapshot of the SSF-based navigation system under testing in CARLA.}
  \label{fig:simulator}
  \vspace{-0.5cm} 
\end{figure}
As depicted in Fig. \ref{fig:testframework}, we setup a SSF-based navigation system testing platform in CARLA \cite{Dosovitskiy17} that is a high-fidelity simulator built on the Unreal Engine. 
Our platform consists of the motion planner module, the ego-vehicle module and the environment module. Communication between the two modules is achieved via ROS.
The motion planner algorithm integrates scene flow information to enhance obstacle avoidance and path prediction. 
Specifically, the environment module handles the physical motion data, and then ego-vehicle module can get the observation. After the scene flow of each object is extracted, each obstacle's velocity $V_{t}^{k}$ is determined by Eq. (\ref{eq:avg_V}). This velocity information, combined with the positions of the dynamic objects calculated by $P_{t}^{k}$, is then fed into the motion planner algorithm to control the ego-vehicle, which can adjust the ego-vehicle’s path, allowing for effective obstacle avoidance and accurate path prediction as the vehicle navigates through the dynamic environment. This scene flow-based navigation can operate without a map, as its localization can be directly obtained from scene flow information.
As illustrated in Fig. \ref{fig:simulator}, the primary objective is to maneuver the vehicle from way point A to way point B, it follows the navigation route, encounters and overtakes an obstacle vehicle due to its higher speed.

\section{Experiments}
% This section validates SSF performance using public datasets and high-fidelity simulations. Subsection A presents static environment segmentation and SSF odometry experiments. Subsection B covers dynamic obstacles segmentation and SSF-based navigation. Subsection C conducts ablation studies on various innovations.

\subsection{Validation of SSF-SLAM on Public Datasets}
\subsubsection{Datasets}
The whole dataset used for training the ASF and validating the SLAM framework was obtained by combining the SUScape-CARLA\footnote{\url{https://suscape.net/datasets/sceneflow}} dataset (including 14,472 frames) and the KITTI\cite{geiger2013vision} dataset (including 2690 frames). Then, the dataset is split into two parts in equal proportion between SUScape-CARLA and KITTI, \textit{i.e.,} the training dataset that comprises 16,352 frames, and the validation dataset that consists of 810 frames, each frame including 8192 points.
According to the range of the number of moving objects in the scene, the processed dataset is divided into two categories, denoted as $D^H$ and $D^T$, in which sampling 100 foreground points is for $D^H$, and 4000 foreground points for $D^T$.
\begin{table*}
  \begin{threeparttable}[c]
\renewcommand\arraystretch{1.2}
\caption{Results of Translational Relative Pose Error (RPE) [m]}%title
\label{tab:RPE}
\centering
\begin{tabular}{C{1.1cm}|C{1.7cm}|C{0.5cm}C{0.8cm}C{0.8cm}|C{0.5cm}C{0.8cm}C{0.8cm}|C{0.8cm}C{0.8cm}C{0.8cm}|C{0.8cm}C{0.8cm}C{0.8cm}}% four columns
\toprule[2pt] %change the first line to \toprule
\multirow{3}{*}&
\multirow{3}{*}&
\multicolumn{6}{c|}{xyz\tnote{\textbf{a}}}& \multicolumn{6}{c}{rpy}\\
\cline{3-14} %change the second line to midrule
% \midrule %change the second line to midrule
Framework&
Method&
\multicolumn{3}{c|}{$D^H$}&
\multicolumn{3}{c|}{$D^T$}&
\multicolumn{3}{c|}{$D^H$}&
\multicolumn{3}{c}{$D^T$}
\\
\cline{3-14} %change the second line to midrule
&& RMSE\tnote{\textbf{b}} & SSE & STD & RMSE & SSE & STD & RMSE & SSE & STD & RMSE & SSE & STD\\
\cline{1-14} %change the second line to midrule
\multirow{3}{*}{A\_LOAM}&
RANSAC\cite{fischler1981random}
& 4.111 & 1352.264 & 2.183 & 5.356& 1377.533 & 2.225 &
0.244 & 4.562 & 0.192 & 0.385 & 5.127 & 0.202\\
&SF
& 0.455 & 16.553 & 0.378 & 0.544& 160.653 & 0.547
& 0.063 & 2.417 & 0.084 & 0.104 & 3.154 & 0.096\\
&SSF(Ours)
& \textbf{0.102} & \textbf{0.447} & \textbf{0.053} & \textbf{0.099}& \textbf{0.432} & \textbf{0.054}
& \textbf{0.034} & \textbf{0.535} & \textbf{0.031} & \textbf{0.019} & \textbf{0.028} & \textbf{0.015}\\
\cline{1-14} %change the second line to midrule
\multirow{3}{*}{S\_LOAM}&
RANSAC\cite{fischler1981random}
& 1.110 & 442.427 & 0.743 & 1.121& 450.674 & 0.844 &
0.230 & 4.228 & 0.168 & 0.232 & 4.237 & 0.175\\
&SF
& 0.322 & 8.278 & 0.261 & 0.488 & 85.555 & 0.351 &
0.073 & 2.427 & 0.064 & 0.083 & 3.135 & 0.104\\
&SSF(Ours)
& \textbf{0.069} & \textbf{0.380} & \textbf{0.043} & \textbf{0.069} & \textbf{0.376} & \textbf{0.047}
& \textbf{0.014} & \textbf{0.016} & \textbf{0.011} & \textbf{0.020} & \textbf{0.032} & \textbf{0.017}\\
\bottomrule[2pt] %change the third line to bottomrule
\end{tabular}
\begin{tablenotes}
  \item [a]
‘xyz’ denotes the translational error, indicating the error in camera position along the three axes (x, y, z).
‘rpy’ represents the rotational error, indicating the error in camera orientation, including the rotations around the camera's fixed coordinate system's x-axis (roll), y-axis (pitch), and z-axis (yaw).
  \item [b]
% \item $MEAN$ represents the average distance between the predicted and ground truth values for each sample point, measured in meters.
RMSE (Root~Mean~Square~Error) represents the average deviation;
SSE (Sum~of~squares~due~to~error) represents the sum of squared errors between predicted and ground truth values for each sample point, measured in meters;
STD (Standard~Deviation) represents the dispersion of the sample data, indicating the distribution of sample data around the mean. 
\end{tablenotes}
  \end{threeparttable}
  \vspace{-0.5cm}
\end{table*}
\subsubsection{Comparative Results}
\begin{figure}[!t]
\centering
\subfloat[]{
		\includegraphics[width=0.9\linewidth]{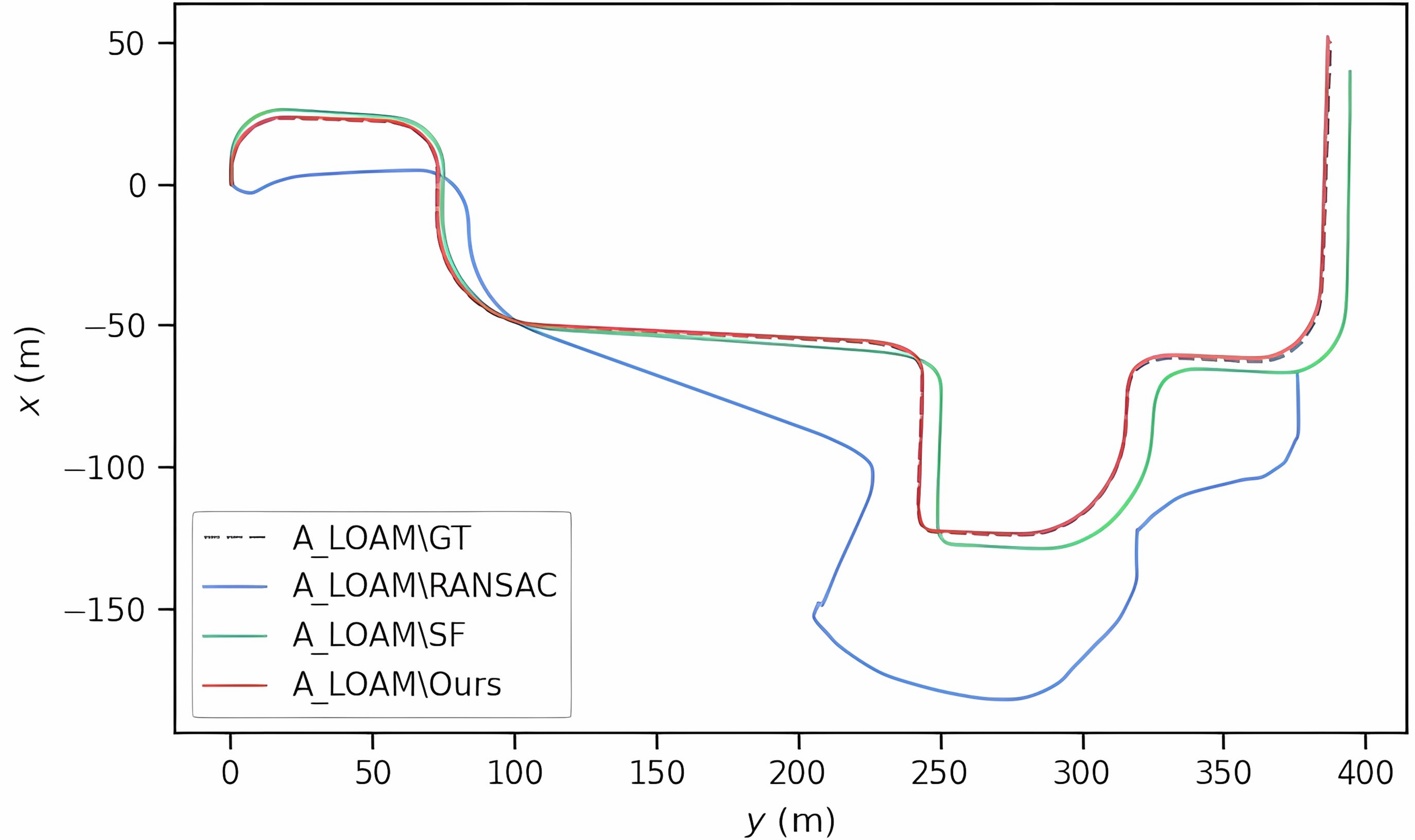}}\\
\subfloat[]{
		\includegraphics[width=0.9\linewidth]{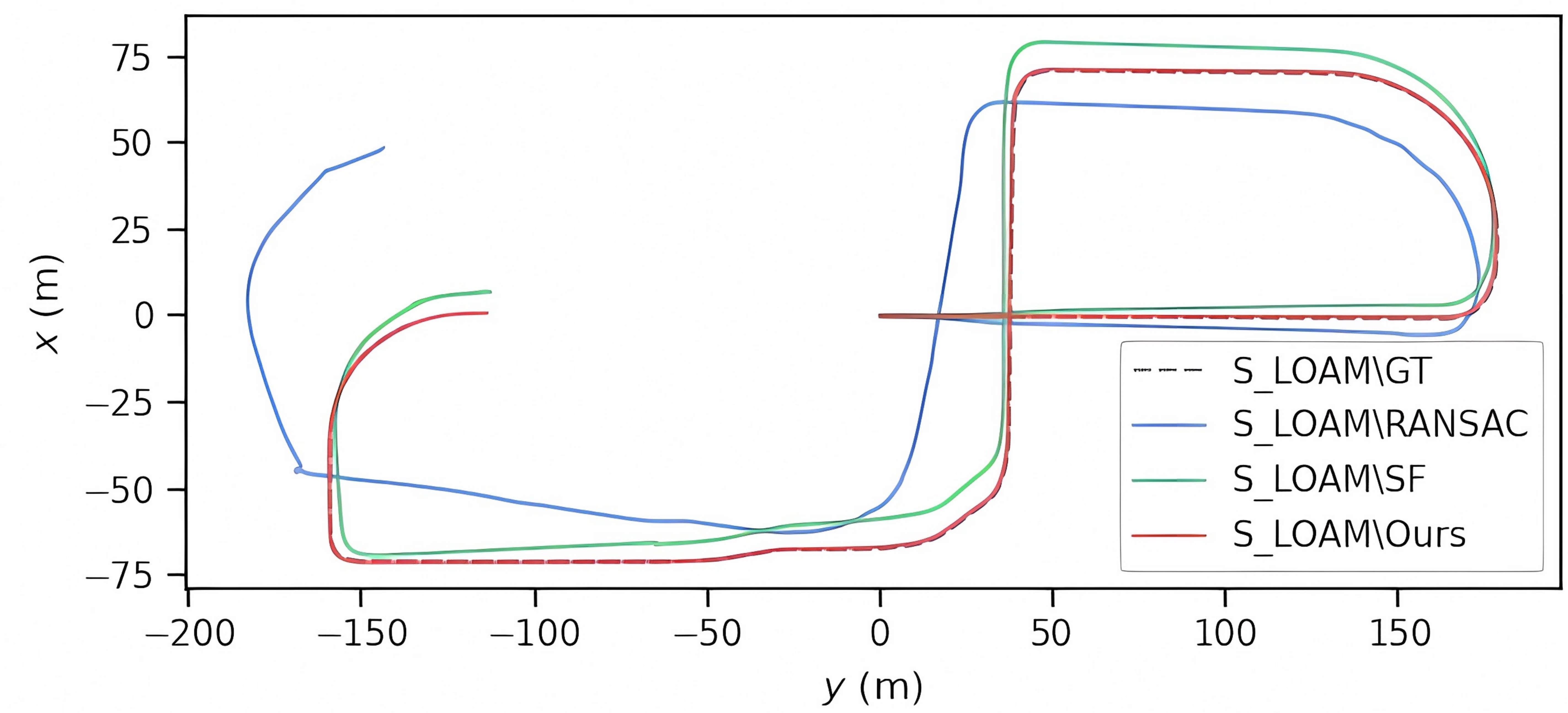}}
\caption{Ego-vehicle trajectory estimation (map construction) results based on the point-cloud SLAM with different odometry modules: (a) A\_LOAM odometry; (b) S\_LOAM odometry.}
\label{fig:SLAM_PATH}
  \vspace{-0.5cm} 
\end{figure}

We conducted a comprehensive assessment of the SLAM system integrated with SSF estimation to validate its localization efficacy and map construction performance in traffic scenarios. Here, two SLAM frameworks are used for experiments, \textit{i.e.}, A\_LOAM\footnote{\url{https://github.com/HKUST-Aerial-Robotics/A-LOAM.git}} and S-LOAM\footnote{\url{https://github.com/haocaichao/S-LOAM.git}}. A-LOAM is an advanced implementation of LOAM\cite{zhang2014loam}, which uses Eigen and Ceres solver to simplify code structure but without IMU. S-LOAM is a LiDAR SLAM framework that, compared with A-LOAM, adds odometry loop closure optimization. The odometry modules of both SLAM frameworks are same, with their initial point cloud registration strategies relying on the statistical RANSAC\cite{fischler1981random} by default. We compare the localization precision in SLAM using the different odometry modules on both $D^H$ and $D^T$ datasets, as shown in Table \ref{tab:RPE}, where SF is that only use the scene flow information for odometry process, and SSF is our proposed method that merges the motion segmentation into the scene flow. It can be seen that the SF achieves a significant fall in trajectory errors on all evaluation metrics. After adding the semantic information, our SSF outperforms with the superior performance, resulting in a reduction of more than 93\%, compared to the RANSAC module. As shown in Fig. \ref{fig:SLAM_PATH}, the experimental results highlight that in dynamic traffic scenarios, the trajectory error of the SLAM system enhanced by SSF estimation is very small compared to the ground truth (GT).
\subsection{Testing Framework for Vehicle Navigation in CARLA}
\subsubsection{Experiments Setting}
\begin{figure}
  \centering 
  % \vspace{-0.5cm} 
  \includegraphics[width=1.05\linewidth]{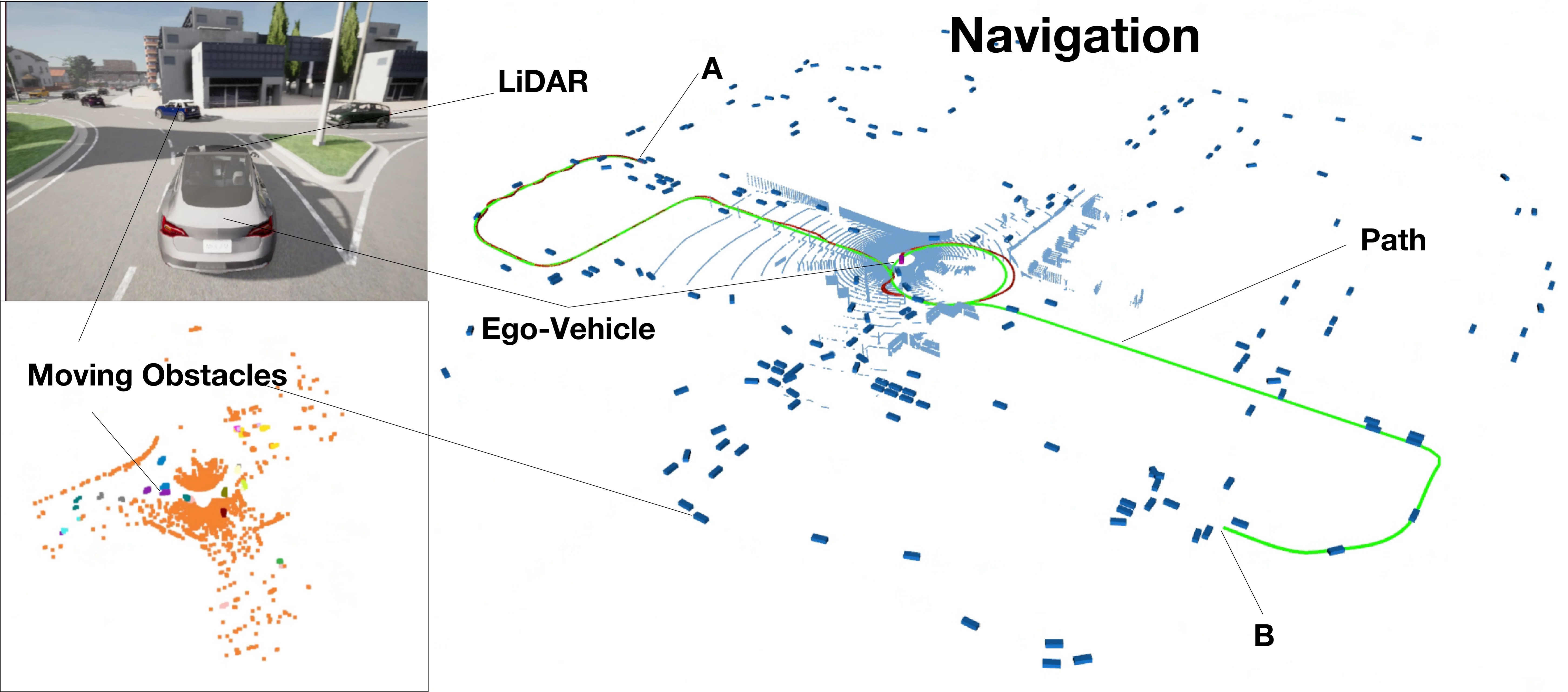}
  \caption{An illustration of SSF-based autonomous navigation in complex traffic scenes in CARLA.}
\label{fig:benchmark}
\vspace{-0.5cm} 
\end{figure}
Navigating vehicles through traffic scenarios using only onboard sensing and computing resources is challenging due to stringent accuracy and latency requirements. We will perform experiments in CARLA\cite{Dosovitskiy17} to demonstrate that SSF is capable for the real-time navigation task. A 64-line 3D LiDAR sensor mounted atop the vehicle is used to provide real-time environmental data. The experimental setup is illustrated in Fig. \ref{fig:benchmark}, where the vehicle, operating in differential steering mode, must traverse between two way points as quickly as possible while avoiding being surrounded by numerous moving obstacles. 
A varying number of vehicles and pedestrians are randomly placed within the test map, exhibiting reciprocal collision avoidance behavior through speed adjustments. The ego vehicle is equipped with a LiDAR sensor to detect obstacles. We use the ground truth of each obstacle's orientation, position, and speed, as well as RDA\cite{10036019}, a state-of-the-art optimization-based dynamic collision avoidance MPC solution, as benchmarks. The calculated poses and geometric information are provided to RDA for planning. We compare our method with DBSCAN\cite{backlund2011density}, a commonly used point cloud-based foreground point instance segmentation scheme, and PointRCNN\cite{Shi_2019_CVPR}. All these methods use a reference speed of 10 m/s, and the path length between two ways on the map is 961 meters.

\subsubsection{Comparative Results}
\begin{table*}
  \begin{threeparttable}[c]
\renewcommand\arraystretch{1.2}
\caption{Quantitative Results for Fig. \ref{fig:system} }%title
\label{tab:ORNavigation}
\centering
\begin{tabular}
{C{1.0cm}|C{0.95cm}C{0.95cm}C{0.95cm}C{0.95cm}|C{0.95cm}C{0.95cm}C{0.95cm}C{0.95cm}|C{0.95cm}C{0.95cm}C{0.95cm}C{0.95cm}}%
\toprule[2pt] %change the first line to \toprule
Moving &
\multicolumn{4}{c|}{Success Rate$\uparrow$}& \multicolumn{4}{c|}{Navigation Time(s)$\downarrow$}& \multicolumn{4}{c}{Moving Average Speed(m/s)$\uparrow$}\\
\cline{2-13} %change the second line to midrule
% \midrule %change the second line to midrule
Obstacles&
{\tiny GT} & {\tiny DBSCAN} & {\tiny PointRCNN} &  {\tiny SSF-PAN} & {\tiny GT} & {\tiny DBSCAN} & {\tiny PointRCNN} & {\tiny SSF-PAN} & {\tiny GT} & {\tiny DBSCAN} & {\tiny PointRCNN} & {\tiny SSF-PAN}\\
\cline{1-13} %change the second line to midrule
50 & 0.96 & 0.88 & 0.91 & 0.92 & 308.42 & 466.28 & 453.74 & 430.63 &
11.11 & 8.89 & 10.23  & 10.28 \\
100 & 0.88 & 0.74 & 0.83  & 0.86 & 353.68& 536.78 & 438.06  & 418.18 &
9.72 & 8.06 & 8.99  & 9.17 \\
200 & 0.80 & 0.63 & 0.68 & 0.76 & 630.73& 690.36 & 679.46 & 677.73 &
9.44 & 7.22 & 7.98 & 8.06 \\
500 & 0.73 & 0.50 & 0.57 & 0.73 & 722.56& 777.65 & 750.98 & 741.47 &
9.17 & 5.94 & 6.03 & 6.94 \\
800 & 0.70 & 0.44 & 0.52 & 0.68 & 739.54& 839.24 & 820.35 & 804.58 &
7.22 & 5.83 & 5.97 & 6.39 \\
1000 & 0.65 & 0.33 & 0.46 & 0.60 & 848.43& 939.57 & 877.35 & 866.73 &
6.67 & 5.56 & 5.65 & 6.29 \\

\bottomrule[2pt] %change the third line to bottomrule
\end{tabular}
  \end{threeparttable}
  \vspace{-0.5cm}
\end{table*}
The evaluation comprises three metrics: success rate, navigation time, and moving average speed. A successful case is defined as the vehicle completing travel between two way points without experiencing any collisions. The success rate is calculated as the proportion of successful cases out of 100 trials. Navigation time refers to the duration taken by the vehicle to complete the route, recorded by the Rviz platform. Therefore, we also compare the moving average speeds of different approaches. Experimental results based on 100 trials with obstacle numbers ranging from 50 to 1,000 are presented in Table \ref{tab:ORNavigation}.

The results show that SSF outperforms DBSCAN and PointRCNN \cite{Shi_2019_CVPR} in terms of success rate, navigation time, and moving average speed. Particularly, when the number of traffic participants exceeds 500, the advantage of SSF becomes more pronounced as obstacle density increases. Although the superiority of SSF over PointRCNN\cite{Shi_2019_CVPR}  is not significant when the obstacle count is below 500, in scenarios requiring SLAM operations, any object detection method must rely on other mapping methods. In contrast, SSF can output the point clouds required for mapping in one step. This allows SSF to provide more accurate speed and pose information for obstacle avoidance algorithms through precise scene flow computation, foreground-background point segmentation, and foreground dynamic object instance segmentation, thereby significantly enhancing overall performance and reliability.
\subsection{Ablation Studies}
\subsubsection{Semantic Segmentation Evaluation}
\begin{table}[!t]
  \centering
% \vspace{-0.5cm} 
  \begin{threeparttable}[c]
  \renewcommand\arraystretch{1.2}
    \caption{Clustering/Semantic Segmentation Accuracy}
    \label{tab:ClusteringSemanticAccuracy}
    \begin{tabular}{C{2.5cm} C{1.5cm} C{1.5cm} C{1.5cm}}
      \toprule
      Segmentation Method/Accuracy (\%) & Only Point Cloud & Only Scene Flow & Point Cloud \& Scene Flow(Ours)                         \\
      \midrule
      OGC\cite{NEURIPS2022_c6e38569}  & 89.79 & 20.54 & -                     \\
      GMM\cite{reynolds2009gaussian}  & 60.38 & 87.33 & 90.25                 \\
      DBSCAN\cite{backlund2011density}          & 42.97/46.62 & OM\tnote{a} / \textbf{85.01}  & 42.96/46.72      \\
      PointNet++\cite{qi2017pointnet++}      & \textbf{87.33} & 30.52 & 85.61                     \\
      SSF(Ours)      & - & - & \textbf{93.27}                     \\
      \bottomrule
    \end{tabular}
    \begin{tablenotes}\tiny
      \item [a] OM stands for Out of Memory. 
    \end{tablenotes}
    \vspace{-0.5cm} 
  \end{threeparttable}
\end{table}
\begin{table*}[!h]
  \centering
  \begin{threeparttable}[c]
    \renewcommand\arraystretch{1.2}
    \caption{Scene Flow Estimation Accuracy Averaged over 200 Scenes}
    \label{tab:Scene_Flow_Prediction_Accuracy}
  \begin{tabular}{C{1.5cm}| C{3.5cm}| C{2.5cm}| C{2.5cm}| C{2.5cm}| C{2.5cm}}
    \toprule
    Dataset & Feature Extraction & EPE3D(m)$\downarrow$ & AS(\%) $\uparrow$ & AR(\%) $\uparrow$ & Outliers(\%)$\downarrow$ \\
    \midrule
	\multirow{3}*{$D^H$}
                 & Only Point Cloud & 0.107400 & 76.8931 & 89.2483 & 69.6648 \\
		       ~ & Implicit Strategy & 0.073223 & 78.4000 & 89.4864 & 65.7234 \\
               ~ & Explicit Strategy  & \textbf{0.039528} & \textbf{92.3349} & \textbf{97.4034} & \textbf{62.2180} \\
\cline{1-6} %change the second line to midrule
    \multirow{3}*{$D^T$}
                 & Only Point Cloud & 0.051007 & 87.1849 & 95.7969 & 63.4523 \\
               ~ & Implicit Strategy & 0.086719 & 75.1837 & 86.8626 & 60.1704 \\
               ~ & Explicit Strategy  & \textbf{0.037604} & \textbf{92.4589} & \textbf{97.5734} & \textbf{62.4444} \\
    \bottomrule
  \end{tabular}
  \end{threeparttable}
    \vspace{-0.5cm} 
\end{table*}

Table~\ref{tab:ClusteringSemanticAccuracy} compares the performance of five different clustering/segmentation methods for dynamic-static scene segmentation in our system. The results show that when point cloud data is concatenated with scene flow data, the segmentation accuracy improves significantly. Specifically, GMM shows a significant 29.97\% increase in accuracy. DBSCAN also benefits from scene flow, with accuracy rising to 46.72\%, though its improvement is limited by point cloud distribution. Notably, DBSCAN's accuracy increases to 85.01\% when segmenting only scene flow. PointNet++, designed primarily for point cloud features, experiences a slight decrease in accuracy when scene flow is added, dropping from 87.33\% to 85.61\%. Our SSF method, based on the modified OGC network, achieves the highest accuracy of 93.27\%, demonstrating its superior performance in complex, dynamic environments.
\subsubsection{Scene Flow Evaluation}
This experiment explores two strategies to achieve the specified objective based on ASF: implicit and explicit approaches.

\textbf{Implicit Strategy:}
This strategy modifies input and output data to achieve implicit enhancements. It sets dynamic objects' scene flow to a distinct values from background motion. This preprocessing step retains semantic information in the point cloud while disregarding dynamic object, focusing on the static motion relative to the observing vehicle. 

\textbf{Explicit Strategy:}
The core idea of this method is to combine semantic information and geometric information from point cloud data as input. It aims to incorporate more semantic context into point cloud data, potentially enhancing the accuracy of scene flow computation.

Table~\ref{tab:Scene_Flow_Prediction_Accuracy} compares these two strategies and only point cloud extracted as features during the scene flow prediction. 
The improvements of the scene flow estimation with the
active scene flow dataset in explicit strategy can achieve at most 0.06 meters in
EPE3D, 15.44\% in AS, and 8.16\% in AR, respectively. 
Therefore, in the final experiment, this study employed the explicit strategy for scene flow prediction.
\begin{figure}
\centering
\subfloat[]{
		\includegraphics[width=0.52\linewidth]{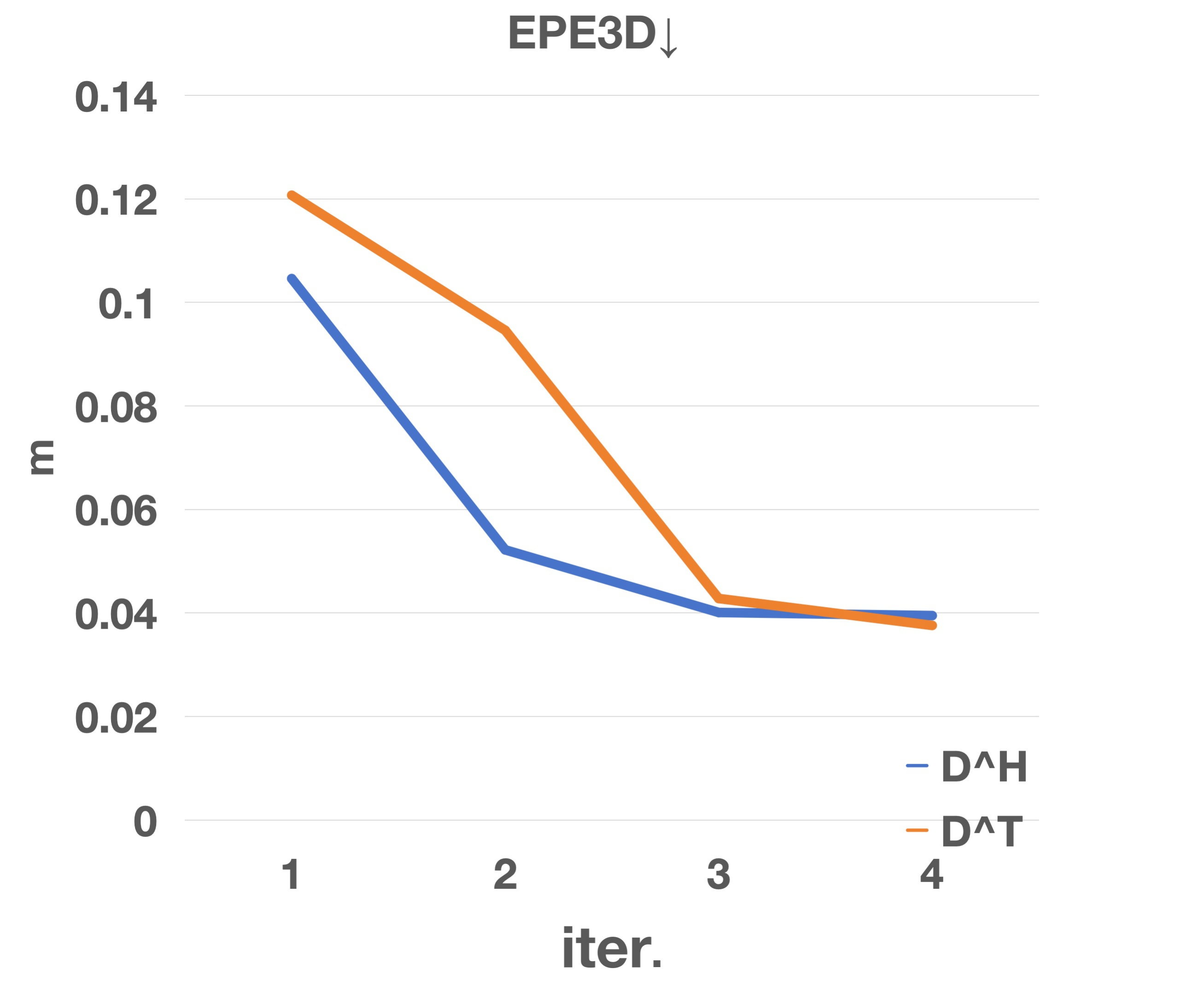}}
\subfloat[]{
		\includegraphics[width=0.52\linewidth]{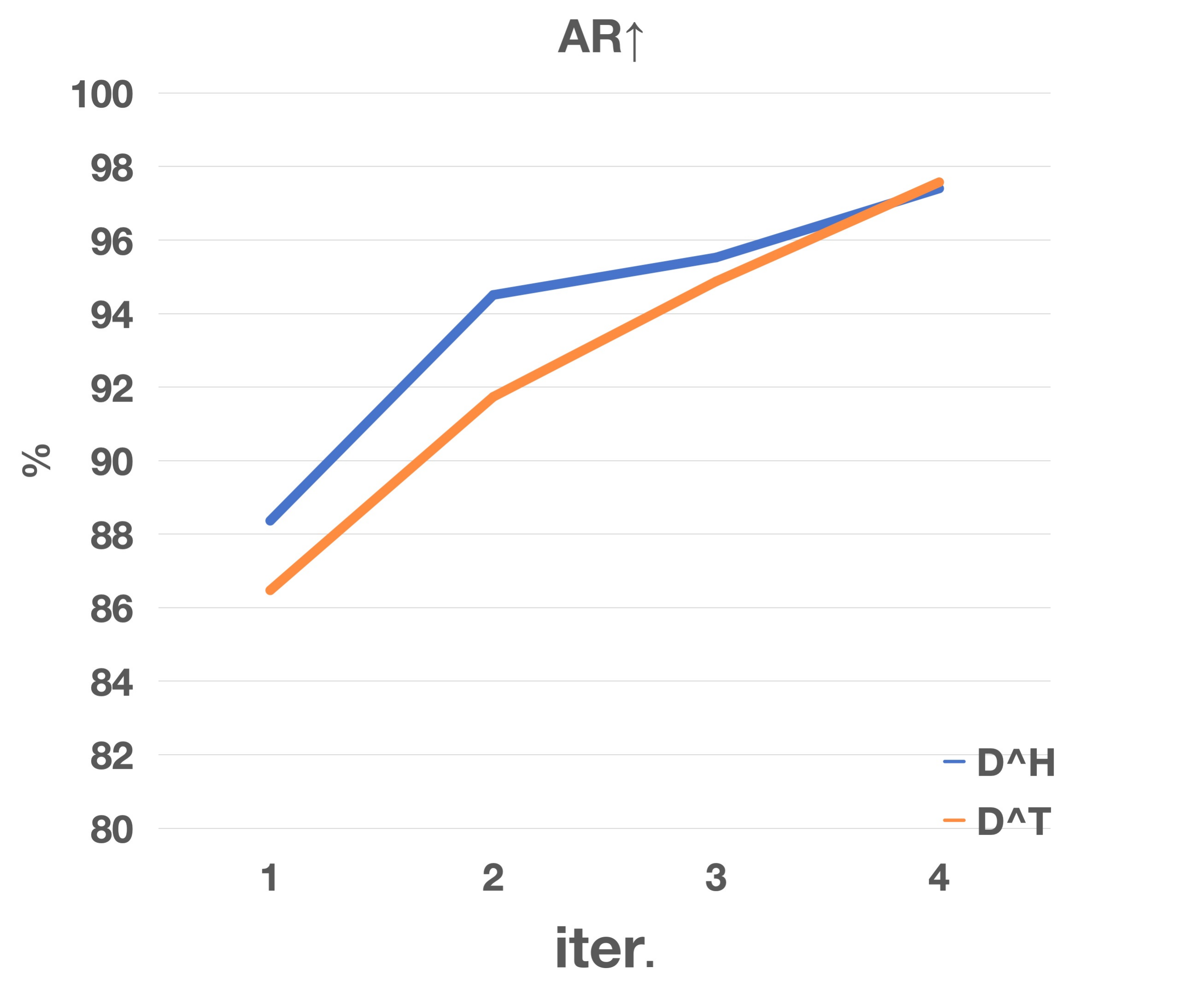}}
\caption{The ablation experiments for the iterative optimization.}
\label{fig:sforesult}
  % \vspace{-0.5cm} 
\end{figure}
\begin{table}[!t]
  \centering
  % \vspace{-0.5cm} 
  \begin{threeparttable}[c]
    \caption{SLAM Trajectory Error for Scene Flow Odometry Module}
    \label{tab:SFAE}
  \begin{tabular}{C{0.3cm} C{0.5cm} C{1.0cm} C{1.5cm} C{0.8cm} C{0.8cm} C{0.8cm}}
    \toprule
    PC & IMU & Our\_Seg & Our\_SF & MEAN\tnote{\textbf{a}}$\thinspace \downarrow$ & RMSE$\downarrow$ & STD$\downarrow$                       \\
    \midrule
    \CheckmarkBold & $\times$ & $\times$ & $\times$ & 25.97& 26.64& 5.94  \\
    \CheckmarkBold & $\times$& $\times$ & \CheckmarkBold& 5.66& 6.35& 2.88  \\
    \CheckmarkBold & \CheckmarkBold &$\times$ & \CheckmarkBold& 3.99& 4.26& 1.51  \\
    \CheckmarkBold & $\times$ & \CheckmarkBold & \CheckmarkBold& \textbf{3.06}& \textbf{3.85}& \textbf{1.33}  \\
    \bottomrule
  \end{tabular}
  \begin{tablenotes}\tiny
  \item [a]
MEAN represents the average distance between the predicted and ground truth values for each sample point, measured in meters.
\end{tablenotes}
  \end{threeparttable}
  \vspace{-0.5cm} 
\end{table}
\subsubsection{Results of Iterative Framework for Scene Flow Estimation and Motion Segmentation }
Regarding the innovative aspects of the mutual promotion network, our ablation study involves a longitudinal comparison where segmentation information and point cloud information are continuously used as inputs to compare the accuracy of scene flow estimation. Finally, the iterative framework of ASF and OGC is adopted. As shown in the Fig. \ref{fig:sforesult}, with the increase in iterations, the error decreases, the accuracy of scene flow estimation improves.
\subsubsection{Scene Flow Odometry Ablation Experiment}
A series of ablation experiments, detailed in Table~\ref{tab:SFAE}, evaluated the impact of using the same raw data set on point cloud SLAM trajectory error computation within a consistent map. The methods compared in the table include: (1) PC: point cloud data as input, (2) IMU: inertial measurement unit aiding odometry calibration, (3) Our\_Seg: OGC and Point Cloud \& Scene Flow used as input for clustering segmentation, and (4) Our\_SF: scene flow optimization used to output scene flow information. Results show that using scene flow and semantic information as odometry significantly reduces SLAM trajectory errors, improving localization accuracy and efficiency, especially with large-scale datasets.

\section{CONCLUSION}
This paper has presented a Semantic Scene Flow-based Perception systems for Autonomous Navigation (SSF-PAN) in complex traffic scenarios, which includes a SSF module, an iterative optimization framework and a testing platform. 
The proposed SSF neural network based on OGC with the improved loss function can enhance the segmentation accuracy of both dynamic objects and the static environment.
The proposed iterative optimization framework can improve both scene flow estimation and segmentation performance. 
The developed SSF-based navigation platform can ensure robust autonomous navigation by continuously assessing the performance of the SSF perception system in simulation environments, enabling map-free navigation in complex traffic scenarios. 
The experimental findings indicate that our SSF method exhibits superior performance in managing complex traffic conditions compared to current techniques across different navigation tasks, with approximately $1\%\sim81\%$ improvement in success rate, $2\% \sim 8\%$ reduction in navigation time, and a $10\% \sim 17\%$ increase in the average moving speed.
Future research will develop a high-precision multi-task model for scene flow estimation and point cloud segmentation without iterative processing.

% \nocite{*}
\bibliographystyle{unsrt}
\bibliography{ref/refs} 

\vfill

\end{document}